\documentclass{article}

\usepackage[preprint]{neurips_2023}

\usepackage{amsmath,amsfonts,bm}

\def\eqref#1{equation~\ref{#1}}

\def\1{\bm{1}}

\DeclareMathAlphabet{\mathsfit}{\encodingdefault}{\sfdefault}{m}{sl}
\SetMathAlphabet{\mathsfit}{bold}{\encodingdefault}{\sfdefault}{bx}{n}

\def\gE{{\mathcal{E}}}

\def\gX{{\mathcal{X}}}

\usepackage[utf8]{inputenc} 
\usepackage[T1]{fontenc}    
\usepackage{hyperref}       
\usepackage{url}            
\usepackage{booktabs}       
\usepackage{amsfonts}       
\usepackage{nicefrac}       
\usepackage{microtype}      
\usepackage[dvipsnames]{xcolor}
\usepackage{cleveref}
\usepackage{graphicx}
\usepackage{subcaption}

\title{Towards equilibrium molecular conformation generation with GFlowNets}

\author{%
Alexandra Volokhova\thanks{Equal contribution.}\\
Mila, Université  de Montréal\\
\And
Michał Koziarski\footnotemark[1]\\
Mila, Université  de Montréal\\
\And
Alex Hernández-García\\
Mila, Université  de Montréal\\
\And
Cheng-Hao Liu\\
Mila, McGill University\\
\And
Santiago Miret\\
Intel Labs\\
\And
Pablo Lemos\\
Mila, Université  de Montréal, \\Ciela Institute, Flatiron Institute\\
\And
Luca Thiede\\
Vector Institute, University of Toronto\\
\And
Zichao Yan\\
Mila, Université  de Montréal\\
\And
Alán Aspuru-Guzik\\
Vector Institute, University of Toronto\\
\And
Yoshua Bengio\\
Mila, Université  de Montréal\\
}

\begin{document}

\maketitle
\begin{abstract}
 Sampling diverse, thermodynamically feasible molecular conformations plays a crucial role in predicting properties of a molecule. In this paper we propose to use GFlowNet for sampling conformations of small molecules from the Boltzmann distribution, as determined by the molecule's energy. The proposed approach can be used in combination with energy estimation methods of different fidelity and discovers a diverse set of low-energy conformations for highly flexible drug-like molecules. We demonstrate that GFlowNet can reproduce molecular potential energy surfaces by sampling proportionally to the Boltzmann distribution.
\end{abstract}

\section{Introduction}

Molecules exist in the three-dimensional space as a distribution of atomic positions, referred to as conformations.
Given the temperature of the system, the probability of each conformation to occur 
is defined
by its formation energy, and follows a Boltzmann distribution~\citep{mcquarrie1997physical}. 
In many computational drug-discovery processes, it is crucial to know the set of most probable, i.e. low-energy, conformations to predict properties of interest~\citep{boehr2009role, trott2010autodockvina}.
In addition, exploring the potential energy surface by sampling proportionally to the Boltzmann distribution can give key chemical insights such as transition pathways and electron transfer~\citep{schlegel2003jcp,benniston2006csr}.

Among computational chemistry methods, molecular dynamics simulation is the standard approach, where methods such as CREST 
have shown feasibility to accurately access numerous low-energy conformations~\citep{prachat2020crest}. However, it remains computationally expensive for high throughput applications and large compounds. Faster alternatives with knowledge-based algorithms, such as distance-geometry methods like ETKDG \citep{riniker2015rdkit}, cannot sample in accordance to the Boltzmann distribution and quickly deteriorate with increasing molecular size.

Machine learning (ML) generative models are promising for conformation generation of molecules \citep{ganea2021geomol, xu2022geodiff, jing2022torsional}. However, they are traditionally focused on maximum likelihood training on a dataset, which does not guarantee sampling proportionally to the Boltzmann distribution. Several recent works such as Boltzmann generators are approaching this problem 
\citep{noe2019boltzmann, kohler2021smooth, jing2022torsional, zheng2023towards}, but none of them has yet demonstrated sufficient generality (see Appendix~\ref{sec:rel-work} for details).

In this paper, we use generative flow networks (GFlowNets) for sampling molecular equillibrium conformations from the Boltzmann distribution. We focus on sampling torsion angles of a molecule, as they contain most of the variance of the conformation space while bond lengths and angles can be efficiently generated by fast rule-based methods.
\citet{gfn_contin} presented a proof of concept to demonstrate the capability of GFlowNet to sample from a distribution defined on a two-dimensional torus. Here, we extend this work to a more realistic setting of an arbitrary number of torsion angles. Furthermore, we train GFlowNets with several energy estimation methods of varying fidelity. We experimentally demonstrate that the proposed approach can sample molecular conformations from the Boltzmann distribution, producing diverse, low-energy conformations for a wide range of drug-like molecules with varying number (2-12) of torsion angles.

\section{GFlowNet for conformation generation}

GFlowNets were originally introduced as a learning algorithm for amortized probabilistic inference in high-dimensional discrete spaces \citep{gfn_original} and a generalisation to continuous or hybrid spaces was recently introduced by \citet{gfn_contin}. The method is designed for sampling from an unnormalised probability density which is often represented as the reward function $R(x)$ over the sample space $x\in\gX$. 

The sampling process starts from a source state $s_0$ and continues with a trajectory of sequential updates $\tau = (s_0\to s_1\to \dots \to s_n = x)$  according to a trainable forward policy $p_F(s_{t} | s_{t-1}; \theta)$, which defines probability of the forward transition $s_{t-1}\to s_t$. Once the termination transition $s_{n-1}\to x$ is sampled, the reward function $R(x)$ provides a signal for computing the training objective. In addition to the trainable forward policy, GFlowNet can learn a backward policy $p_B(s_{t-1} | s_{t}; \theta)$ for modelling the probability of the backward transitions $s_{t}\to s_{t-1}$. It gives additional flexibility to the forward policy $p_F$ allowing to model a rich family of distributions over $\gX$.

In this paper, we propose the use of GFlowNets as a generative model to sample molecular conformations in the space of torsion angles. 
Formally, we can describe the space of $d$ torsion angles of a molecule as the hyper-torus defined in $\mathcal{X} = [0, 2\pi)^d$. To sample trajectories $\tau$ with GFlowNets that satisfy the theoretical assumptions defined by \citep{gfn_contin}, we establish a fixed number of steps $T$ and include the step number into the state. This yields a state space $\mathcal{S} = \{s_0\} \cup [0, 2\pi)^d \times \{1, 2 \ldots T\}$. 
In order to learn expressive distributions on the hyper-tori, we parameterized independently the forward and backward policies $p_F$ and $p_B$ and trained the GFlowNets with the Trajectory Balance objective~\citep{gfn_tb} (\Cref{sec:apx-tb}). Both forward and backward policies are parameterized with a multilayer perceptron (MLP) which outputs parameters for a mixture of von Mises distribution. Note that in this setting we need to train an individual GFlowNet for every molecule as different molecules may have different numbers of torsion angles.

In order to sample from the Boltzmann distribution, we define the reward function using the energy of the molecular conformation in the following way:

\begin{equation}
R(x) =  e^{-\gE(C(x))\beta},
\end{equation}
where $\gE(C(x))$ is the potential energy of the molecular conformation $C(x)$ in vacuum and $C(x)$ is defined by the sampled torsion angles $x$. Other parameters of the conformation $C$ such as bond lengths and angles are sampled with ETKDG \citep{riniker2015rdkit} and fixed during the GFlowNet training. The positive scalar $\beta$ corresponds to the inverse temperature of the molecule, but we treat it as a hyper-parameter of the method for the scope of this work. 

The energy function is computed as an approximation of quantum mechanical density-functional theory (DFT) and we consider several estimators of different fidelity in our experiments. We use the molecule's potential energy in vacuum for the scope of our work, but the method can be applied to any energy function of interest.

\section{Empirical evaluation}

We conducted an experimental study aimed at addressing several research questions: whether the approach can sample conformers proportionally to the Boltzmann distribution, its capacity to generate diverse low-energy conformations, and how the choice of the energy estimator impacts performance. Additionally, we examined how the proposed approach scales with an increasing number of torsion angles. Firstly, we conducted experiments on molecules with only two considered torsion angles to perform a more in-depth analysis of the results which was not possible in higher dimensions. Then, we scaled up our study, considering a broader range of molecules with varying numbers of torsion angles. Training details can be found in Appendix~\ref{sec:exp-params}.

\subsection{Set-up}

Conformation generation is conditioned on an input molecular graph encoded as a SMILES string. We first process SMILES with RDKit library \citep{landrum2016rdkit} which uses ETKDG to generate an initial conformation. This defines bond lengths and angles and non-rotatable torsion angles (see Appendix \ref{sec:non-rotatable-tas} for more details). Then, GFlowNet generates rotatable torsion angles and their values are updated accordingly in the conformation.

\paragraph{Energy estimation} 
We experimented with a representative set the existing approaches to energy estimation with different trade-offs between accuracy and computational costs. The most accurate of the considered methods is a semiempirical quantum chemical method, GFN2-xTB \citep{bannwarth2019gfn2}, which is designed for estimating energies of molecular systems for accurate conformation generation. 
We also employ a faster and less precise force-field approach, GFN-FF \citep{bannwarth2021extended}. 
Finally, we consider TorchANI \citep{gao2020torchani}, which implements a neural network potential called ANI \citep{devereux2020extending} for energy estimation of organic molecules. Its computational cost is comparable to GFN-FF, but applicability is limited to the specific domain of molecules it was trained on. 

\paragraph{Data}
In the two-dimensional setting, we used alanine dipeptide, ibuprofen and ketorolac molecules. All of them have two main torsion angles largely affecting their energy.
For experiments with multiple torsion angles, we used molecules from the GEOM-DRUGS dataset \citep{axelrod2022geom}, a popular benchmark containing low-energy conformations of 304k drug-like molecules, which were generated with meta-dynamics simulation of potential energy estimated by GFN2-xTB \citep{prachat2020crest}. The average number of rotatable torsion angles for GEOM-DRUGS molecules is $7.9$ and $92.8\%$ of them have less than $13$ rotatable torsion angles (\Cref{sec:apx-geom}).

\subsection{Two-dimensional setting}

\begin{figure}[t]
    \centering
    \begin{subfigure}{0.25\textwidth}
        \includegraphics[width=\linewidth]{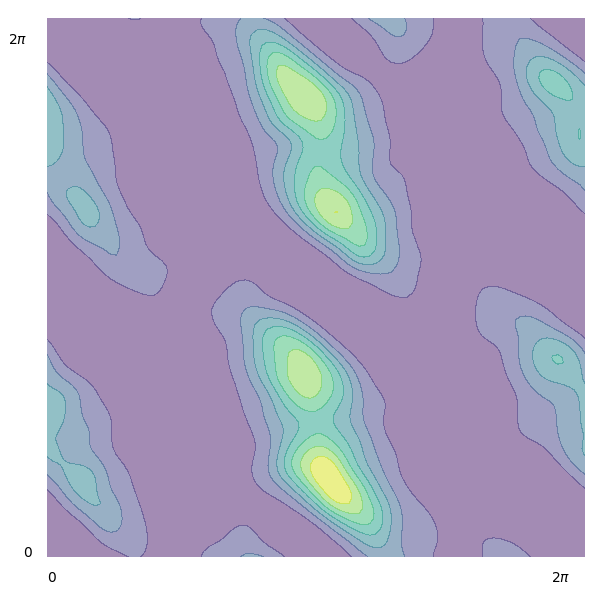}
        \caption{Reward GFN2-xTB}
        \label{fig:ketorolac-reward-xtb}
    \end{subfigure}
    \begin{subfigure}{0.25\textwidth}
        \includegraphics[width=\linewidth]{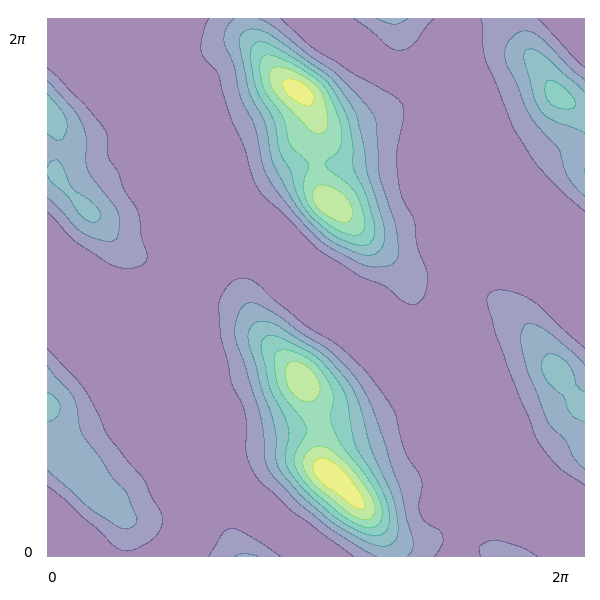}
        \caption{GFlowNet GFN2-xTB}
        \label{fig:ketorolac-gfn-xtb}
    \end{subfigure}
    \begin{subfigure}{0.25\textwidth}
        \includegraphics[width=\linewidth]{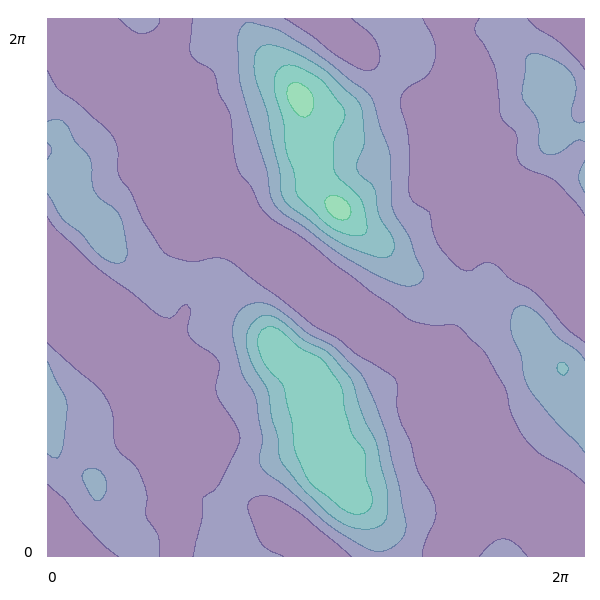}
        \caption{MCMC GFN2-xTB}
        \label{fig:ketorolac-mcmc-xtb}
    \end{subfigure}
    \\
    \begin{subfigure}{0.25\textwidth}
        \includegraphics[width=\linewidth]{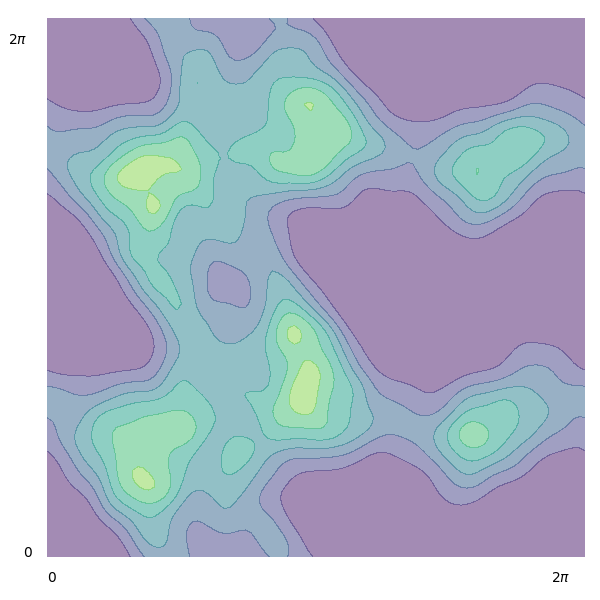}
        \caption{Reward TorchANI}
        \label{fig:ibuprofen-reward-torchani}
    \end{subfigure}
    \begin{subfigure}{0.25\textwidth}
        \includegraphics[width=\linewidth]{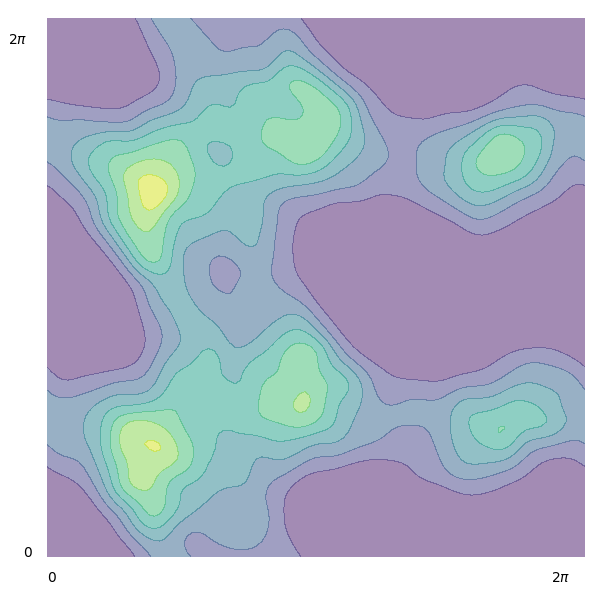}
        \caption{GFlowNet TorchANI}
        \label{fig:ibuprofen-gfn-torchani}
    \end{subfigure}
    \begin{subfigure}{0.25\textwidth}
        \includegraphics[width=\linewidth]{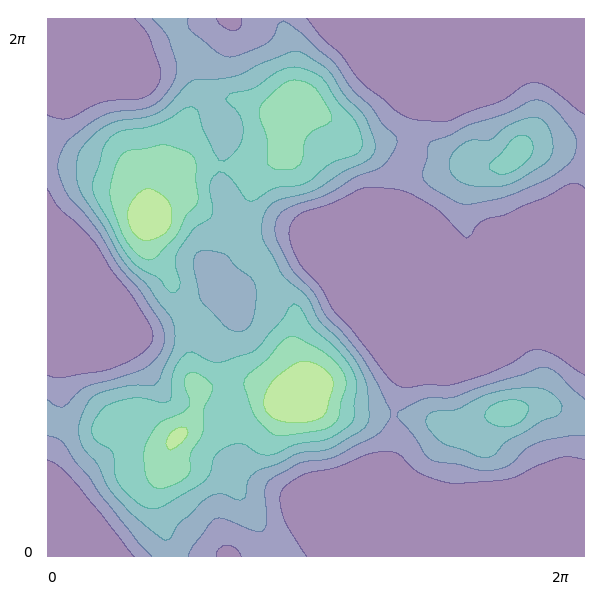}
        \caption{MCMC TorchANI}
        \label{fig:ibuprofen-mcmc-torchani}
    \end{subfigure}
    \caption{KDE on samples from the reward function (left), GFlowNet (centre) and MCMC (right) for two molecules and two different proxies: ketorolac from GFN2-xTB (top) and ibuprofen from TorchANI (bottom).}
    \label{fig:kde}
\end{figure}

We begin by investigating the performance of GFlowNet in simple, well-studied molecular systems in two dimensions: alanine dipeptide, ibuprofen, and ketorolac~\citep{vargas2002adp, zeng2023qdpi}. 
In this experiment, we aim to assess how well the proposed approach can learn to sample from the target distribution and analyze the impact of the energy estimator.

The low dimensionality allows us to visualize the kernel density estimation and evaluate the performance numerically using the Jensen–Shannon divergence (JSD). We used nested sampling \citep{buchner2023nested} to produce reference ground-truth energy surfaces, and compared the proposed approach with MCMC as an example of an unamortized method for sampling from an unnormalized probability distribution.

In Figure~\ref{fig:kde}, we present the obtained potential energy surfaces of ketorolac and ibuprofen. Comparable analysis for all three molecules and proxies, as well as the computed JSD values, can be found in Appendix~\ref{sec:2d-apx}. As can be seen, while the choice of the energy estimator can influence the overall shape of the energy surface, both considered methods accurately reproduce the ground truth energy surface for all estimators. 
Interestingly, GFlowNet outperformed MCMC in some cases, producing energy surfaces more closely resembling the ground truth.

\subsection{Scaling to multiple torsion angles}




\begin{figure}[t]
    \centering
    \includegraphics[width=\textwidth]{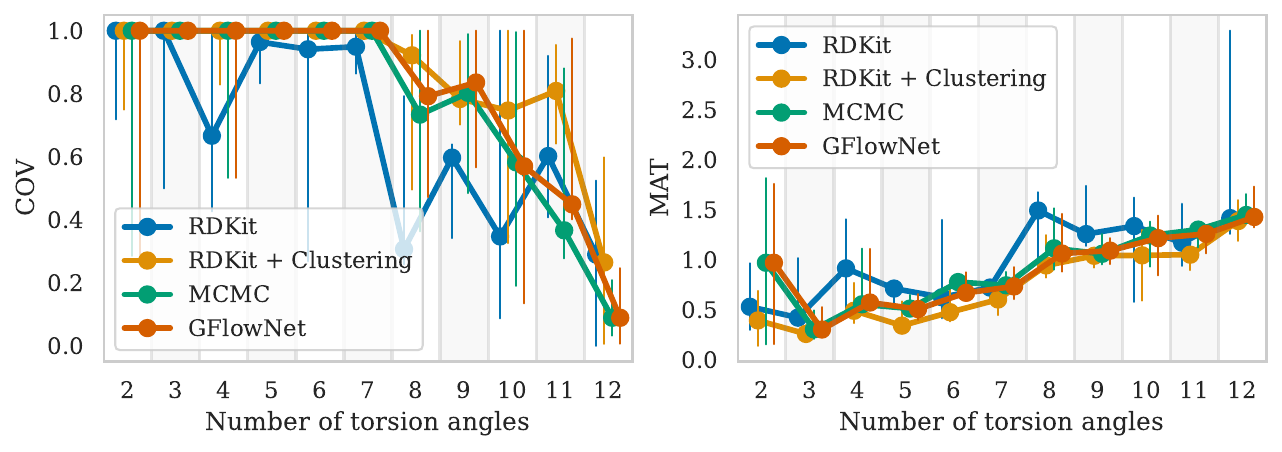}
    \caption{COV and MAT metrics for samples from GFlowNet, RDKit, RDKit+Clustering, MCMC. The markers indicate the median out of five molecules and the vertical lines the range between minimum and maximum values.}
    \label{fig:covmat}
\end{figure}

\begin{figure}[t]
    \centering
    \includegraphics[width=0.7\textwidth]{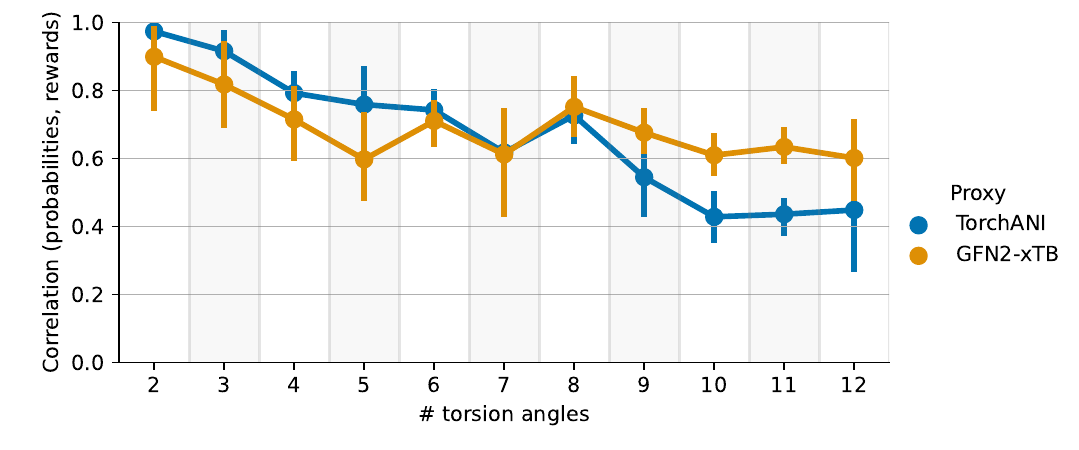}
    \caption{Correlation between estimated sample probabilty and its reward, as a function of the number of torsion angles.}
    \label{fig:corr}
\end{figure}

We then examined a setting with a higher number of torsion angles, up to 12. For each number of torsion angles we selected 5 different molecules from the GEOM-Drugs dataset, and trained GFlowNet using TorchANI energy estimator separately for each one of them. We compared our method with several baselines: MCMC, ETKDG \citep{riniker2015rdkit} implemented by the RDKit library, as well as a recent approach combining ETKDG with clustering \citep{zhou2023deep}, which was shown to outperform most existing machine learning methods in the low energy conformation generation task. Note that both ETKDG-based approaches perform a qualitatively different task of generating optimized, low-energy conformers, whereas GFlowNet and MCMC aim at sampling from the underlying Boltzmann distribution. In this experiment, we investigate whether diverse low-energy conformations are present among the GFlowNet samples.

In \cref{fig:covmat}, we report the COV and MAT metrics (Appendix~\ref{sec:covmat}), which evaluate the proportion of the reference conformations among samples (COV) and the proximity of the sampled conformations to the reference onces (MAT). We use a set of 1000 generated conformations for all methods to compute the metrics.
As can be seen, GFlowNet outperforms vanilla RDKit, and achieves performance comparable to MCMC and RDKit-clustering, indicating that it can sample diverse low-energy conformations that closely match the ground-truth dataset even when using low-fidelity energy estimator (TorchANI).

Finally, to provide additional evaluation of GFlowNet’s scalability to higher dimensions, we examined the correlation between the estimated probability to sample a conformation with a GFlowNet (Appendix~\ref{sec:prob-est}) and its energy. We performed this analysis for GFlowNet exclusively because other methods do not provide a straightforward way for estimating sample probability. For the sake of this comparison, we also considered GFN2-xTB as an energy estimator. The results are presented in Figure~\ref{fig:corr}. As can be seen, while the correlation declines with the increasing number of torsion angles, it remains relatively high throughout the considered range, serving as a further indication that the proposed approach can scale to high dimensions. Interestingly, the higher correlation in higher dimensions with GFN2-xTB may suggest that it is easier to learn than TorchANI, probably because of its more accurate modelling of the energy surface.

\section{Discussion and future work}


In this paper, we have proposed a method for sampling molecular conformations using GFlowNet, which samples proportionally to the Boltzmann distribution. We experimentally evaluated the proposed approach and demonstrated that it can sample diverse, low-energy conformations, can be used in combination with energy estimators of different fidelity, and scales well to a higher number of torsion angles.

While the proposed method still requires training individually for every molecule, the described work can be treated as a stepping stone towards developing a generalised GFlowNet model, which could be trained on a whole data set of molecules and sample torsion angles for an arbitrary molecule conditioned on its molecular graph. This can potentially allow us to amortize the computational costs of sampling proportionally to the Boltzmann distribution compared to methods such as MCMC. 


\begin{ack}
We thank Léna Néhale Ezzine and Prudencio Tossou for their contributions to our discussions. We thank Will Hua for his contribution in the code development at the initial stages of the project. We acknowledge computational support from Intel Labs and Mila compute resources for necessary experiments. We acknowledge Mila's IDT team support. 
\end{ack}

\bibliography{references}

\begin{thebibliography}{40}
\providecommand{\natexlab}[1]{#1}
\providecommand{\url}[1]{\texttt{#1}}
\expandafter\ifx\csname urlstyle\endcsname\relax
  \providecommand{\doi}[1]{doi: #1}\else
  \providecommand{\doi}{doi: \begingroup \urlstyle{rm}\Url}\fi

\bibitem[Arts et~al.(2023)Arts, Satorras, Huang, Zuegner, Federici, Clementi, No{\'e}, Pinsler, and Berg]{arts2023two}
Marloes Arts, Victor~Garcia Satorras, Chin-Wei Huang, Daniel Zuegner, Marco Federici, Cecilia Clementi, Frank No{\'e}, Robert Pinsler, and Rianne van~den Berg.
\newblock Two for one: Diffusion models and force fields for coarse-grained molecular dynamics.
\newblock \emph{arXiv preprint arXiv:2302.00600}, 2023.

\bibitem[Axelrod \& Gomez-Bombarelli(2022)Axelrod and Gomez-Bombarelli]{axelrod2022geom}
Simon Axelrod and Rafael Gomez-Bombarelli.
\newblock {GEOM}, energy-annotated molecular conformations for property prediction and molecular generation.
\newblock \emph{Scientific Data}, 9\penalty0 (1):\penalty0 185, 2022.

\bibitem[Bannwarth et~al.(2019)Bannwarth, Ehlert, and Grimme]{bannwarth2019gfn2}
Christoph Bannwarth, Sebastian Ehlert, and Stefan Grimme.
\newblock Gfn2-xtb—an accurate and broadly parametrized self-consistent tight-binding quantum chemical method with multipole electrostatics and density-dependent dispersion contributions.
\newblock \emph{Journal of chemical theory and computation}, 15\penalty0 (3):\penalty0 1652--1671, 2019.

\bibitem[Bannwarth et~al.(2021)Bannwarth, Caldeweyher, Ehlert, Hansen, Pracht, Seibert, Spicher, and Grimme]{bannwarth2021extended}
Christoph Bannwarth, Eike Caldeweyher, Sebastian Ehlert, Andreas Hansen, Philipp Pracht, Jakob Seibert, Sebastian Spicher, and Stefan Grimme.
\newblock Extended tight-binding quantum chemistry methods.
\newblock \emph{Wiley Interdisciplinary Reviews: Computational Molecular Science}, 11\penalty0 (2):\penalty0 e1493, 2021.

\bibitem[Bengio et~al.(2021)Bengio, Jain, Korablyov, Precup, and Bengio]{gfn_original}
Emmanuel Bengio, Moksh Jain, Maksym Korablyov, Doina Precup, and Yoshua Bengio.
\newblock Flow network based generative models for non-iterative diverse candidate generation.
\newblock \emph{Advances in Neural Information Processing Systems}, 34:\penalty0 27381--27394, 2021.

\bibitem[Benniston \& Harriman(2006)Benniston and Harriman]{benniston2006csr}
Andrew~C. Benniston and Anthony Harriman.
\newblock Charge on the move: how electron-transfer dynamics depend on molecular conformation.
\newblock \emph{Chem. Soc. Rev.}, 35:\penalty0 169--179, 2006.
\newblock \doi{10.1039/B503169A}.
\newblock URL \url{http://dx.doi.org/10.1039/B503169A}.

\bibitem[Boehr et~al.(2009)Boehr, Nussinov, and Wright]{boehr2009role}
David~D Boehr, Ruth Nussinov, and Peter~E Wright.
\newblock The role of dynamic conformational ensembles in biomolecular recognition.
\newblock \emph{Nat. Chem. Bio.}, 5\penalty0 (11):\penalty0 789--796, 2009.

\bibitem[Buchner(2023)]{buchner2023nested}
Johannes Buchner.
\newblock Nested sampling methods.
\newblock \emph{Statistic Surveys}, 17:\penalty0 169--215, 2023.

\bibitem[Devereux et~al.(2020)Devereux, Smith, Huddleston, Barros, Zubatyuk, Isayev, and Roitberg]{devereux2020extending}
Christian Devereux, Justin~S Smith, Kate~K Huddleston, Kipton Barros, Roman Zubatyuk, Olexandr Isayev, and Adrian~E Roitberg.
\newblock Extending the applicability of the ani deep learning molecular potential to sulfur and halogens.
\newblock \emph{Journal of Chemical Theory and Computation}, 16\penalty0 (7):\penalty0 4192--4202, 2020.

\bibitem[Fu et~al.(2022)Fu, Wu, Wang, Xie, Keten, Gomez-Bombarelli, and Jaakkola]{fu2022forces}
Xiang Fu, Zhenghao Wu, Wujie Wang, Tian Xie, Sinan Keten, Rafael Gomez-Bombarelli, and Tommi Jaakkola.
\newblock Forces are not enough: Benchmark and critical evaluation for machine learning force fields with molecular simulations.
\newblock \emph{arXiv preprint arXiv:2210.07237}, 2022.

\bibitem[Ganea et~al.(2021)Ganea, Pattanaik, Coley, Barzilay, Jensen, Green, and Jaakkola]{ganea2021geomol}
Octavian Ganea, Lagnajit Pattanaik, Connor Coley, Regina Barzilay, Klavs Jensen, William Green, and Tommi Jaakkola.
\newblock Geomol: Torsional geometric generation of molecular 3d conformer ensembles.
\newblock \emph{Advances in Neural Information Processing Systems}, 34:\penalty0 13757--13769, 2021.

\bibitem[Gao et~al.(2020)Gao, Ramezanghorbani, Isayev, Smith, and Roitberg]{gao2020torchani}
Xiang Gao, Farhad Ramezanghorbani, Olexandr Isayev, Justin~S Smith, and Adrian~E Roitberg.
\newblock Torchani: A free and open source pytorch-based deep learning implementation of the ani neural network potentials.
\newblock \emph{Journal of chemical information and modeling}, 60\penalty0 (7):\penalty0 3408--3415, 2020.

\bibitem[Gelman \& Rubin(1992)Gelman and Rubin]{gelman1992inference}
Andrew Gelman and Donald~B Rubin.
\newblock Inference from iterative simulation using multiple sequences.
\newblock \emph{Statistical science}, 7\penalty0 (4):\penalty0 457--472, 1992.

\bibitem[Gogineni et~al.(2020)Gogineni, Xu, Punzalan, Jiang, Kammeraad, Tewari, and Zimmerman]{gogineni2020torsionnet}
Tarun Gogineni, Ziping Xu, Exequiel Punzalan, Runxuan Jiang, Joshua Kammeraad, Ambuj Tewari, and Paul Zimmerman.
\newblock Torsionnet: A reinforcement learning approach to sequential conformer search.
\newblock \emph{Advances in Neural Information Processing Systems}, 33:\penalty0 20142--20153, 2020.

\bibitem[Hawkins et~al.(2010)Hawkins, Skillman, Warren, Ellingson, and Stahl]{hawkins2010conformer_omega}
Paul~CD Hawkins, A~Geoffrey Skillman, Gregory~L Warren, Benjamin~A Ellingson, and Matthew~T Stahl.
\newblock Conformer generation with omega: algorithm and validation using high quality structures from the protein databank and cambridge structural database.
\newblock \emph{Journal of chemical information and modeling}, 50\penalty0 (4):\penalty0 572--584, 2010.

\bibitem[Jing et~al.(2022)Jing, Corso, Chang, Barzilay, and Jaakkola]{jing2022torsional}
Bowen Jing, Gabriele Corso, Jeffrey Chang, Regina Barzilay, and Tommi Jaakkola.
\newblock Torsional diffusion for molecular conformer generation.
\newblock \emph{Advances in Neural Information Processing Systems}, 35:\penalty0 24240--24253, 2022.

\bibitem[K{\"o}hler et~al.(2021)K{\"o}hler, Kr{\"a}mer, and No{\'e}]{kohler2021smooth}
Jonas K{\"o}hler, Andreas Kr{\"a}mer, and Frank No{\'e}.
\newblock Smooth normalizing flows.
\newblock \emph{Advances in Neural Information Processing Systems}, 34:\penalty0 2796--2809, 2021.

\bibitem[Lahlou et~al.(2023)Lahlou, Deleu, Lemos, Zhang, Volokhova, Hern{\'a}ndez-Garc{\i}a, Ezzine, Bengio, and Malkin]{gfn_contin}
Salem Lahlou, Tristan Deleu, Pablo Lemos, Dinghuai Zhang, Alexandra Volokhova, Alex Hern{\'a}ndez-Garc{\i}a, L{\'e}na~N{\'e}hale Ezzine, Yoshua Bengio, and Nikolay Malkin.
\newblock A theory of continuous generative flow networks.
\newblock In \emph{International Conference on Machine Learning}, pp.\  18269--18300. PMLR, 2023.

\bibitem[Landrum(2016)]{landrum2016rdkit}
G~Landrum.
\newblock Rdkit: open-source cheminformatics http://www. rdkit. org.
\newblock \emph{Google Scholar There is no corresponding record for this reference}, 3\penalty0 (8), 2016.

\bibitem[Malkin et~al.(2022)Malkin, Jain, Bengio, Sun, and Bengio]{gfn_tb}
Nikolay Malkin, Moksh Jain, Emmanuel Bengio, Chen Sun, and Yoshua Bengio.
\newblock Trajectory balance: Improved credit assignment in gflownets.
\newblock \emph{Advances in Neural Information Processing Systems}, 35:\penalty0 5955--5967, 2022.

\bibitem[McQuarrie \& Simon(1997)McQuarrie and Simon]{mcquarrie1997physical}
Donald~A. McQuarrie and John~D. Simon.
\newblock \emph{Physical Chemistry: A Molecular Approach}.
\newblock University Science Books, Sausalito, CA, 1997.
\newblock Chapter 17.

\bibitem[No{\'e} et~al.(2019)No{\'e}, Olsson, K{\"o}hler, and Wu]{noe2019boltzmann}
Frank No{\'e}, Simon Olsson, Jonas K{\"o}hler, and Hao Wu.
\newblock Boltzmann generators: Sampling equilibrium states of many-body systems with deep learning.
\newblock \emph{Science}, 365\penalty0 (6457):\penalty0 eaaw1147, 2019.

\bibitem[Patel \& Tewari(2022)Patel and Tewari]{patel2022rl}
Yash Patel and Ambuj Tewari.
\newblock Rl boltzmann generators for conformer generation in data-sparse environments.
\newblock \emph{arXiv preprint arXiv:2211.10771}, 2022.

\bibitem[Pracht et~al.(2020{\natexlab{a}})Pracht, Bohle, and Grimme]{crest2020}
Philipp Pracht, Fabian Bohle, and Stefan Grimme.
\newblock Automated exploration of the low-energy chemical space with fast quantum chemical methods.
\newblock \emph{Physical Chemistry Chemical Physics}, 22\penalty0 (14):\penalty0 7169--7192, 2020{\natexlab{a}}.

\bibitem[Pracht et~al.(2020{\natexlab{b}})Pracht, Bohle, and Grimme]{prachat2020crest}
Philipp Pracht, Fabian Bohle, and Stefan Grimme.
\newblock Automated exploration of the low-energy chemical space with fast quantum chemical methods.
\newblock \emph{Phys. Chem. Chem. Phys.}, 22:\penalty0 7169--7192, 2020{\natexlab{b}}.
\newblock \doi{10.1039/C9CP06869D}.
\newblock URL \url{http://dx.doi.org/10.1039/C9CP06869D}.

\bibitem[Riniker \& Landrum(2015)Riniker and Landrum]{riniker2015rdkit}
Sereina Riniker and Gregory~A Landrum.
\newblock Better informed distance geometry: using what we know to improve conformation generation.
\newblock \emph{Journal of chemical information and modeling}, 55\penalty0 (12):\penalty0 2562--2574, 2015.

\bibitem[Schlegel(2003)]{schlegel2003jcp}
H.~Bernhard Schlegel.
\newblock Exploring potential energy surfaces for chemical reactions: An overview of some practical methods.
\newblock \emph{Journal of Computational Chemistry}, 24\penalty0 (12):\penalty0 1514--1527, 2003.
\newblock \doi{https://doi.org/10.1002/jcc.10231}.
\newblock URL \url{https://onlinelibrary.wiley.com/doi/abs/10.1002/jcc.10231}.

\bibitem[Schulman et~al.(2017)Schulman, Wolski, Dhariwal, Radford, and Klimov]{ppo}
John Schulman, Filip Wolski, Prafulla Dhariwal, Alec Radford, and Oleg Klimov.
\newblock Proximal policy optimization algorithms.
\newblock \emph{CoRR}, abs/1707.06347, 2017.

\bibitem[Shi et~al.(2021)Shi, Luo, Xu, and Tang]{shi2021learning}
Chence Shi, Shitong Luo, Minkai Xu, and Jian Tang.
\newblock Learning gradient fields for molecular conformation generation.
\newblock In \emph{International conference on machine learning}, pp.\  9558--9568. PMLR, 2021.

\bibitem[Th{\"o}lke \& De~Fabritiis(2021)Th{\"o}lke and De~Fabritiis]{tholke2021equivariant}
Philipp Th{\"o}lke and Gianni De~Fabritiis.
\newblock Equivariant transformers for neural network based molecular potentials.
\newblock In \emph{International Conference on Learning Representations}, 2021.

\bibitem[Torrado \& Lewis(2019)Torrado and Lewis]{torrado2019cobaya}
Jes{\'u}s Torrado and Antony Lewis.
\newblock Cobaya: Bayesian analysis in cosmology.
\newblock \emph{Astrophysics Source Code Library}, pp.\  ascl--1910, 2019.

\bibitem[Torrado \& Lewis(2021)Torrado and Lewis]{torrado2021cobaya}
Jesus Torrado and Antony Lewis.
\newblock Cobaya: Code for bayesian analysis of hierarchical physical models.
\newblock \emph{Journal of Cosmology and Astroparticle Physics}, 2021\penalty0 (05):\penalty0 057, 2021.

\bibitem[Trott \& Olson(2010)Trott and Olson]{trott2010autodockvina}
Oleg Trott and Arthur~J. Olson.
\newblock Autodock vina: Improving the speed and accuracy of docking with a new scoring function, efficient optimization, and multithreading.
\newblock \emph{Journal of Computational Chemistry}, 31\penalty0 (2):\penalty0 455--461, 2010.
\newblock \doi{https://doi.org/10.1002/jcc.21334}.
\newblock URL \url{https://onlinelibrary.wiley.com/doi/abs/10.1002/jcc.21334}.

\bibitem[Vargas et~al.(2002)Vargas, Garza, Hay, and Dixon]{vargas2002adp}
Rubicelia Vargas, Jorge Garza, Benjamin~P. Hay, and David~A. Dixon.
\newblock Conformational study of the alanine dipeptide at the mp2 and dft levels.
\newblock \emph{The Journal of Physical Chemistry A}, 106\penalty0 (13):\penalty0 3213--3218, 2002.
\newblock \doi{10.1021/jp013952f}.
\newblock URL \url{https://doi.org/10.1021/jp013952f}.

\bibitem[Wang et~al.(2019)Wang, Olsson, Wehmeyer, P{\'e}rez, Charron, De~Fabritiis, No{\'e}, and Clementi]{wang2019machine}
Jiang Wang, Simon Olsson, Christoph Wehmeyer, Adri{\`a} P{\'e}rez, Nicholas~E Charron, Gianni De~Fabritiis, Frank No{\'e}, and Cecilia Clementi.
\newblock Machine learning of coarse-grained molecular dynamics force fields.
\newblock \emph{ACS central science}, 5\penalty0 (5):\penalty0 755--767, 2019.

\bibitem[Xu et~al.(2021)Xu, Luo, Bengio, Peng, and Tang]{xu2021learning}
Minkai Xu, Shitong Luo, Yoshua Bengio, Jian Peng, and Jian Tang.
\newblock Learning neural generative dynamics for molecular conformation generation.
\newblock \emph{arXiv preprint arXiv:2102.10240}, 2021.

\bibitem[Xu et~al.(2022)Xu, Yu, Song, Shi, Ermon, and Tang]{xu2022geodiff}
Minkai Xu, Lantao Yu, Yang Song, Chence Shi, Stefano Ermon, and Jian Tang.
\newblock Geodiff: A geometric diffusion model for molecular conformation generation.
\newblock \emph{arXiv preprint arXiv:2203.02923}, 2022.

\bibitem[Zeng et~al.(2023)Zeng, Tao, Giese, and York]{zeng2023qdpi}
Jinzhe Zeng, Yujun Tao, Timothy~J. Giese, and Darrin~M. York.
\newblock Qd$\pi$: A quantum deep potential interaction model for drug discovery.
\newblock \emph{Journal of Chemical Theory and Computation}, 19\penalty0 (4):\penalty0 1261--1275, 2023.
\newblock \doi{10.1021/acs.jctc.2c01172}.
\newblock URL \url{https://doi.org/10.1021/acs.jctc.2c01172}.
\newblock PMID: 36696673.

\bibitem[Zheng et~al.(2023)Zheng, He, Liu, Shi, Lu, Feng, Ju, Wang, Zhu, Min, et~al.]{zheng2023towards}
Shuxin Zheng, Jiyan He, Chang Liu, Yu~Shi, Ziheng Lu, Weitao Feng, Fusong Ju, Jiaxi Wang, Jianwei Zhu, Yaosen Min, et~al.
\newblock Towards predicting equilibrium distributions for molecular systems with deep learning.
\newblock \emph{arXiv preprint arXiv:2306.05445}, 2023.

\bibitem[Zhou et~al.(2023)Zhou, Gao, Wei, Zheng, and Ke]{zhou2023deep}
Gengmo Zhou, Zhifeng Gao, Zhewei Wei, Hang Zheng, and Guolin Ke.
\newblock Do deep learning methods really perform better in molecular conformation generation?
\newblock \emph{arXiv preprint arXiv:2302.07061}, 2023.

\end{thebibliography}
\bibliographystyle{iclr2024_conference}

\appendix

\section{Related works}
\label{sec:rel-work}
\subsection{Conventional approaches }
The most accurate way to get a set of low-energy conformers is based on molecular-dynamics (MD) simulation \citep{crest2020}, which is a computational method for studying the time evolution of physical systems. This method integrates a Newtonian equation of motion with forces obtained by differentiating the potential energy function of the system and employs metadynamics to discover multiple minima of the energy landscape. However, this accuracy comes with high computational costs, despite simplifications of the energy function, making this method not suitable for high-throughput applications and large molecules \citep{axelrod2022geom}. 

Alternatively, cheminformatics methods are a fast and popular approach for conformer generation. They utilize structures from experimental reference datasets, chemical rules and heuristics to generate plausible 3D structures given a molecular graph. While significantly faster than MD simulations, these methods tend to lack accuracy and generalization. ETKDG is the most widely used cheminformatics method for conformer generation \cite{riniker2015rdkit} implemented in the open-source library RDKit and OMEGA \cite{hawkins2010conformer_omega} is a popular commercial software implementing cheminformatics methods.
 
\subsection{Machine learning and reinforcement learning approaches}
Several machine learning approaches have been developed for conformer generation. Recent advancements include GeoMol (\cite{ganea2021geomol}), GeoDiff (\cite{xu2022geodiff}), and Torsional Diffusion (\cite{jing2022torsional}). These methods show good results on the popular benchmark of drug-like molecules GEOM (\cite{axelrod2022geom}), demonstrating a decent performance and allowing for faster generation than MD simulations. However, a recently proposed simple clustering algorithm on top of the conformations generated by RDKit outperforms many of these approaches \citep{zhou2023deep}.
It is important to note that these machine learning methods are mainly trained for maximizing the likelihood of conformations present in the training dataset and therefore not suitable for sampling proportionally to the Boltzmann distribution.
In contrast, the GFlowNet framework utilizes information about the Boltzmann weights of the conformations by querying the reward function and allows for generating conformations proportionally to the Boltzmann distribution.

Another promising avenue in the realm of conformer generation is the application of machine learning to model the force fields (\cite{wang2019machine, arts2023two, tholke2021equivariant, shi2021learning}). This approach involves training a machine learning model for predicting forces in the system.  Then, molecular dynamics equations are unrolled to sample conformations from the local minima of the energy function based on the forces (gradients of the energy) provided by the model. Machine learning models allow for faster computation of the forces giving a speed advantage compared to MD with an energy function based on the first principles. However, molecular dynamics integration with estimated force fields presents significant stability challenges due to error accumulation during integration \cite{fu2022forces}, which limits the applicability of this approach.

Reinforcement learning (RL) is another class of machine learning algorithms which cast the generation of molecular structures into a Markov decision process. Well-known examples include TorsionNet~\citep{gogineni2020torsionnet} which proceeds by sequentially altering the molecular conformation via updating all torsion angles at every step. The algorithm employs proximal policy optimization~\citep{ppo} and learns by querying a force-fields energy estimation. While TorsionNet and other RL methods do not rely on existing molecular conformation dataset — in a setup similar to our method, they lack the theoretical guarantee of exploring the broader ensemble space of molecular conformation, and indeed, \citet{patel2022rl} shows that they fail at recovering a diverse set of conformations. 

\subsection{Boltzmann generators}

A closely relevant direction of work is Boltzmann generators (\cite{noe2019boltzmann, kohler2021smooth}). These are machine learning models aimed at generating samples from the Boltzmann distribution. However, they are based on normalising flows \cite{noe2019boltzmann, kohler2021smooth}, which are known to have limited expressivity and require training a separate model for each molecular graph 
The torsional diffusion model \cite{jing2022torsional} incorporated annealed importance sampling into the training of the diffusion model which allowed for training the model to sample from the target distribution. However, importance sampling brings additional variance into the gradients and becomes challenging with higher dimensions which limits applicability of this approach. Recent work by \cite{zheng2023towards} also attempted to make diffusion models sample from the Boltzmann distribution. They pretrained the score model, forcing it to both follow the Fokker-Plank equation and match the energy gradient
and then trained it on the dataset with the standard maximum-likelihood objective. This approach allows to incorporate of information about energies into the model but does not guarantee sampling from the Boltzmann distribution. 

\section{Trajectory Balance loss}
\label{sec:apx-tb}

Formally, Trajectory Balance loss was defined by \citep{gfn_tb} as:
\begin{equation}
\label{eq:tb}
L_{TB}(\tau; \theta) = \left(\log \frac{Z_\theta \prod_{t=1}^n p_{F}(s_{t} | s_{t-1}; \theta)}{R(s_n) \prod_{t=1}^{n} p_{B}(s_{t-1} | s_{t}; \theta)}  \right)^2.
\end{equation}

In \cref{eq:tb}, $Z_\theta$ is the (trainable) partition function and $R(s_n)$ is the reward function, evaluated on the terminating states of the trajectories, $s_n = x \in \mathcal{X}$.

\section{Non-rotatable torsion angles}
\label{sec:non-rotatable-tas}
Torsion angles between the triplets of adjacent bonds defines the shape of the molecule together with bond lengths and angles. Due to the chemical constraints, some torsion angles do not vary between different conformations and can be accurately identified from a SMILES string by rule-based methods together with the bond lengths and angles. Specifically, torsion angles corresponding to non-single bonds and torsion angles within the rings are considered as non-rotatable.
In our study we kept these torsion angles fixed and did not sample them with the GFlowNet.  

\section{Experimental details}
\label{sec:exp-params}

\subsection{Energy computation}

Due to perceived energy artifacts observed for GFN-FF, we performed energy normalization and clamping for all of the considered energy estimation methods. Specifically, for every molecule we computed 10,000 random conformations, based on them estimated the minimum and maximum energy values, and normalized all of the estimated energy values to a $[0; 1]$ range. Furthermore, to remove outliers we clamped the values between 1st and 99th quantile of the energy values observed in the initial random sample.

\subsection{GFlowNet training}

GFlowNet used separate MLP for forward and backward policy, both consisting of 5 hidden layers with 512 neurons each, and using leaky ReLU activation function. Policy input was using positional encoding, with every torsion angle encoded by 10 different values using trigonometric functions. Policy output consisted of predicted von Mises distribution parameters, with 5 separate components per torsion angle. Trajectories of length 5 were used, meaning that every sample was constructed by making 5 separate steps, sampled from the predicted distribution parameters. Minimum concentration of 4 was used for the von Mises distribution. GFlowNet was trained using Boltzmann reward function with $\beta = 32$. Training of a single molecule lasted 40,000 iterations, and used Adam optimizer with learning rate of 0.0001 for the policy and 0.01 for $Z_\theta$. Batches consisted of 80 on-policy trajectories with probability of random sampling equal to 0.1, and 20 trajectories from replay buffer of best trajectories seen so far, with a total buffer capacity of 1000.

\section{GEOM-DRUGS torsion angle distribution}
\label{sec:apx-geom}

\begin{figure}
    \centering
    \includegraphics[width=0.7\textwidth]{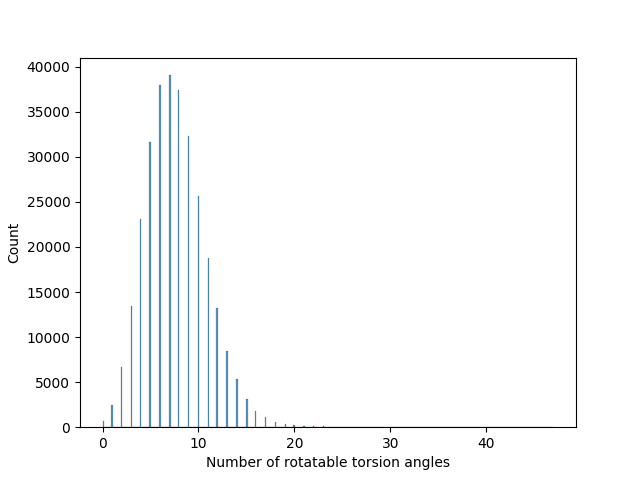}
    \caption{Distribution of torsion angles for molecules in GEOM-DRUGS.}
    \label{fig:dist-geom}
\end{figure}

One important consideration for the experimental study was the range of evaluated torsion angles. To chose an appropriate range we analysed the distribution of torsion angles for molecules contained in GEOM-DRUGS, as presented in \Cref{fig:dist-geom}. As can be seen, the considered range of 2 to 12 torsion angles faithfully represents the dataset, with majority of molecules contained in it.

\subsection{MCMC sampling}

For the Markov-Chain Monte Carlo (MCMC) experiments, we run a Metropolis-Hastings algorithm, with a self-tuning covariance matrix for the proposal step. We use the publicly available {\tt Cobaya} implementation \cite{torrado2019cobaya, torrado2021cobaya}. We use 4 randomly initialized walkers, and run until reaching a Gelman-Rubin~\citep{gelman1992inference} statistic $R-1 < 0.01$. We then burn-in the first $20 \%$ of the chains, and randomly select $1000$ samples from the remaining set.

\section{Evaluation}

\subsection{COVMAT metrics}
\label{sec:covmat}

Root Mean Square Deviation (RMSD) is a metric for evaluating the difference between 3D positions of two conformations, typically computed after first aligning the molecules using a function $\Phi$. Formally, for two molecules $\boldsymbol{R}$, $\hat{\boldsymbol{R}}$ it can be defined as:

\begin{equation}
\operatorname{RMSD}(\boldsymbol{R}, \hat{\boldsymbol{R}})=\min _{\Phi}\left(\frac{1}{n} \sum_{i=1}^n\left\|\Phi\left(\boldsymbol{R}_i\right)-\hat{\boldsymbol{R}}_i\right\|^2\right)^{\frac{1}{2}}
\end{equation}

Coverage (COV) and matching (MAT) are two metrics incorporating RMSD that, given a set of reference low-energy conformations $S_r$, measure to what extent generated samples $S_g$ cover the modes in the reference dataset (COV), and how closely do they resemble specific reference conformations (MAT):

\begin{equation}
\operatorname{COV}\left(S_g, S_r\right)=\frac{\left|\left\{\boldsymbol{R} \in S_r \mid \operatorname{RMSD}(\boldsymbol{R}, \hat{\boldsymbol{R}})<\delta, \hat{\boldsymbol{R}} \in S_g\right\}\right|}{\left|S_r\right|},
\end{equation}

\begin{equation}
\operatorname{MAT}\left(S_g, S_r\right)=\frac{1}{\left|S_r\right|} \sum_{\boldsymbol{R} \in S_r} \min _{\hat{\boldsymbol{R}} \in S_g} \operatorname{RMSD}(\boldsymbol{R}, \hat{\boldsymbol{R}}).
\end{equation}

Note that COVMAT metrics were traditionally used in the task of low-energy molecular conformation generation, with the assumption that generated conformations $S_g$ are close to local energy optima. This does not hold when sampling from Boltzmann distribution, and can produce unfavourable results. Because of that we adjusted the procedure by sampling more conformations than typically seen in the literature, as described in the main text.

We follow the literature \citep{xu2021learning,shi2021learning,zhou2023deep} and set $\delta$ to 1.25\r{A} in our experiments.

\section{Estimation of the log-likelihood of sampling a terminating state $x$}
\label{sec:prob-est}

The log-probability of sampling a terminating state $x$ according to the GFlowNet policy $\pi(x)$ can be expressed as follows:

\[\log \pi(x) = \log \int_{\tau: s_{|\tau|-1}\rightarrow x \in \tau} \prod_{t=0}^{|\tau|-1} P_{F}(s_{t+1}|s_{t};\theta) \,d\tau = \log \int_{\tau: x \in \tau} P_F(\tau) \,d\tau.\]

Note that we are abusing notation in favour of readability in the domain of the integral of the right-most equation, which should be identical to the previous integral's. While computing the integral (or even the corresponding sum, if the state space is discrete but very large) in the above expression is intractable in general, we can efficiently estimate it with importance sampling, by using the backward transition probability distribution $P_B(\tau|x)$ as the importance proposal distribution. Let us recall the core aspects of importance sampling, which is a kind of Monte Carlo simulation method.

Monte Carlo simulation methods can be used to estimate integrals $I(f) = \int_{\tau: x \in \tau} f(\tau)p(\tau) \,d\tau$, as well as very large sums, by drawing $N$ independent and identically distributed (i.i.d.) samples $\{\tau^{(i)}\}_{i=1}^{N}$ and computing an estimate $I_N(f) = \frac{1}{N} \sum_{i=1}^{N}f(\tau^{(i)})$ which converges to $I(f)$ as $N \rightarrow \infty$. Importance sampling introduces an importance proposal distribution $q(\tau)$ whose support includes the support of the target distribution $p(\tau)$, such that we can express the integral $I(f)$ as an expectation over the proposal distribution:

\[I(f) = \int_{\tau: x \in \tau} f(\tau)\frac{p(\tau)}{q(\tau)}q(\tau) \,d\tau = \mathbb{E}_{q(\tau)} [f(\tau)\frac{p(\tau)}{q(\tau)}]\]

By drawing $N$ i.i.d. samples from the proposal distribution $q(\tau)$ we can compute the estimate

\[\hat{I}_N(f) = \frac{1}{N} \sum_{i=1}^{N}f(\tau^{(i)})\frac{p(\tau^{(i)})}{q(\tau^{(i)})}.\]

Returning to the original problem of estimating the log-probability of sampling $x$ with a GFlowNet, we have that the target density is the forward transition probability distribution ($p(\tau) = P_F(\tau)$), the importance proposal distribution is the backward transition probability distribution given $x$ ($q(\tau) = P_B(\tau|x)$), and the sample performance is simply one ($f(\tau) = 1$). Therefore:

\begin{equation}
  \log \hat{\pi}(x) = \log \frac{1}{N} \sum_{i=1}^{N}\frac{P_F(\tau^{(i)})}{P_B(\tau^{(i)}|x)} = \log \sum_{i=1}^{N}\frac{P_F(\tau^{(i)})}{P_B(\tau^{(i)}|x)} - \log N.
\end{equation}

\clearpage
\section{Two-dimensional results}
\label{sec:2d-apx}

\begin{table}[!htb]
    \centering
    \caption{Comparison of JSD for model molecules in the two-dimensional setting.}
    \label{tab:ketorolac}
    \begin{tabular}{llccc}
        \toprule
         Method & Proxy & Alanine Dipeptide& Ibuprofen& Ketorolac\\
         \midrule
         MCMC & TorchANI & \textbf{0.0056} & \textbf{0.0055} & 0.0172\\
          & GFN-FF & \textbf{0.0076} & \textbf{0.0175} & 0.0123\\
          & GFN2-xTB & 0.0090 & \textbf{0.0074} & 0.0143\\
         \midrule
         GFlowNet & TorchANI & 0.0077 & \textbf{0.0055} & \textbf{0.0070}\\
          & GFN-FF & 0.0166 & 0.0182 & \textbf{0.0078}\\
          & GFN2-xTB & \textbf{0.0073} & 0.0075 & \textbf{0.0059}\\
         \bottomrule
    \end{tabular}
\end{table}

\subsection{Ketorolac}

\begin{figure}[!htb]
    \centering
    \begin{subfigure}{0.25\textwidth}
        \includegraphics[width=\linewidth]{images/kde_ketorolac_Reward_xtb.png}
        \caption{Reward GFN2-xTB}
        \label{fig:sup:ketorolac-reward-xtb}
    \end{subfigure}
    \begin{subfigure}{0.25\textwidth}
        \includegraphics[width=\linewidth]{images/kde_ketorolac_GFlowNet_xtb.png}
        \caption{GFlowNet GFN2-xTB}
        \label{fig:sup:ketorolac-gfn-xtb}
    \end{subfigure}
    \begin{subfigure}{0.25\textwidth}
        \includegraphics[width=\linewidth]{images/kde_ketorolac_MCMC_xtb.png}
        \caption{MCMC GFN2-xTB}
        \label{fig:sup:ketorolac-mcmc-xtb}
    \end{subfigure}
    \\
    \begin{subfigure}{0.25\textwidth}
        \includegraphics[width=\linewidth]{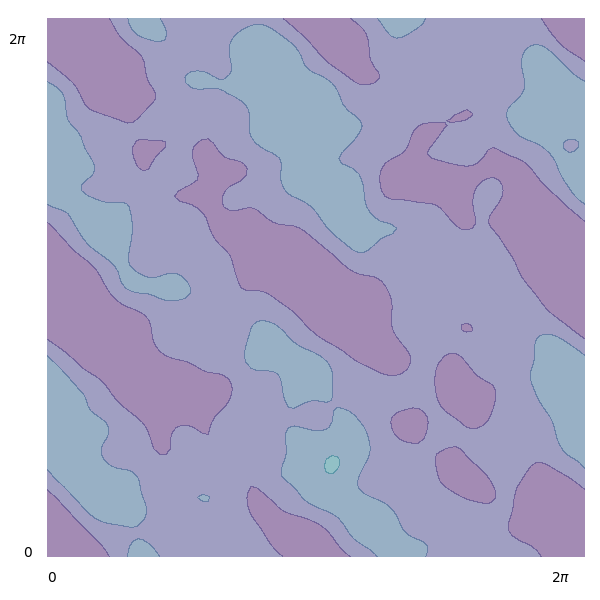}
        \caption{Reward GFN-FF}
        \label{fig:sup:ketorolac-reward-gfn-ff}
    \end{subfigure}
    \begin{subfigure}{0.25\textwidth}
        \includegraphics[width=\linewidth]{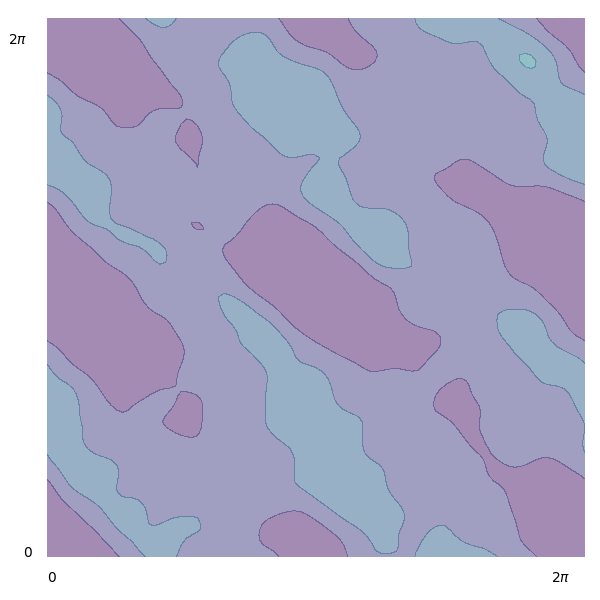}
        \caption{GFlowNet GFN-FF}
        \label{fig:sup:ketorolac-gfn-gfn-ff}
    \end{subfigure}
    \begin{subfigure}{0.25\textwidth}
        \includegraphics[width=\linewidth]{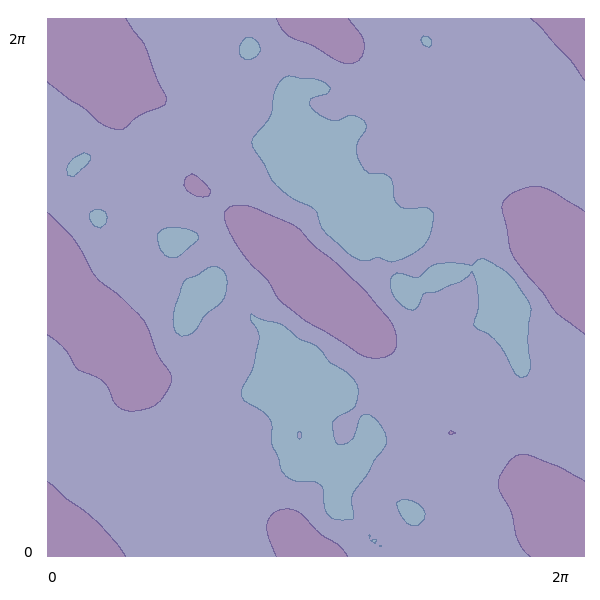}
        \caption{MCMC GFN-FF}
        \label{fig:sup:ketorolac-mcmc-gfn-ff}
    \end{subfigure}
    \\
    \begin{subfigure}{0.25\textwidth}
        \includegraphics[width=\linewidth]{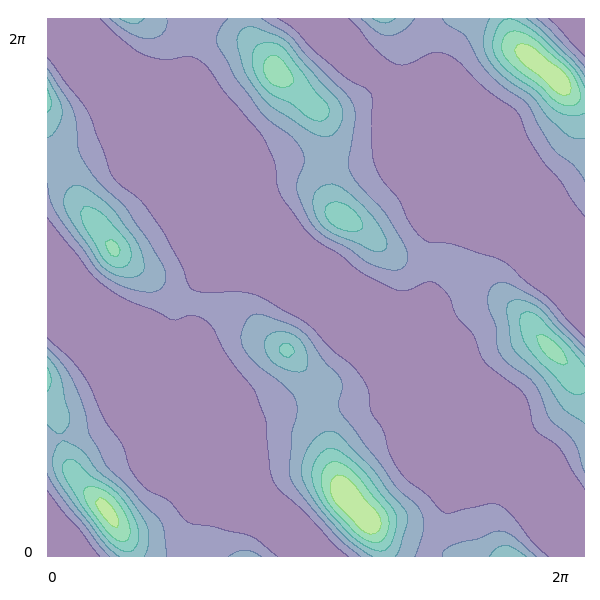}
        \caption{Reward TorchANI}
        \label{fig:sup:ketorolac-reward-torchani}
    \end{subfigure}
    \begin{subfigure}{0.25\textwidth}
        \includegraphics[width=\linewidth]{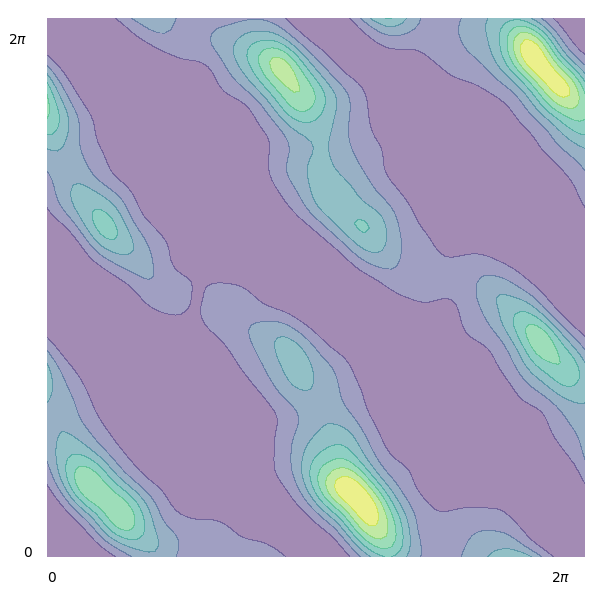}
        \caption{GFlowNet TorchANI}
        \label{fig:sup:ketorolac-gfn-torchani}
    \end{subfigure}
    \begin{subfigure}{0.25\textwidth}
        \includegraphics[width=\linewidth]{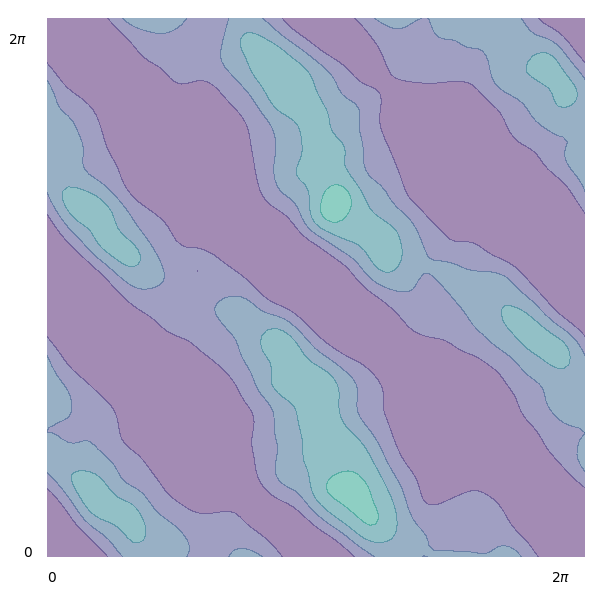}
        \caption{MCMC TorchANI}
        \label{fig:sup:ketorolac-mcmc-torchani}
    \end{subfigure}
    \caption{KDE on samples from the reward function (left) GFlowNet (centre) and MCMC (right) for ketorolac for the three proxies: GFN2-xTB (top), GFN-FF (middle) and TorchANI (bottom).}
\end{figure}

\clearpage
\subsection{Ibuprofen}

\begin{figure}[!htb]
    \centering
    \begin{subfigure}{0.25\textwidth}
        \includegraphics[width=\linewidth]{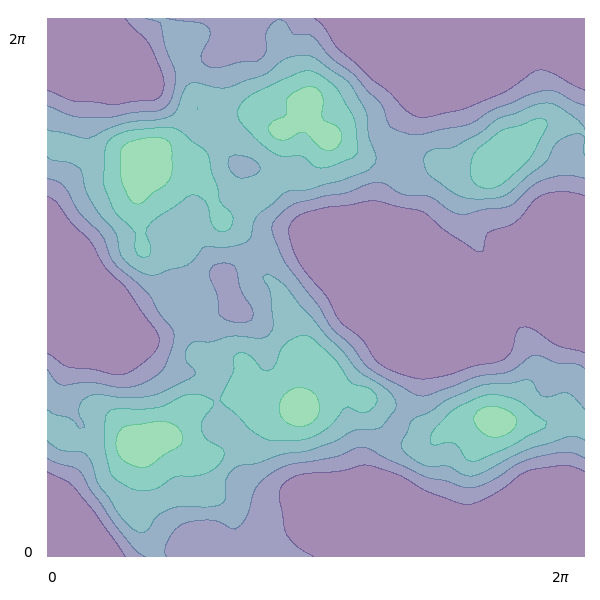}
        \caption{Reward GFN2-xTB}
        \label{fig:sup:ibuprofen-reward-xtb}
    \end{subfigure}
    \begin{subfigure}{0.25\textwidth}
        \includegraphics[width=\linewidth]{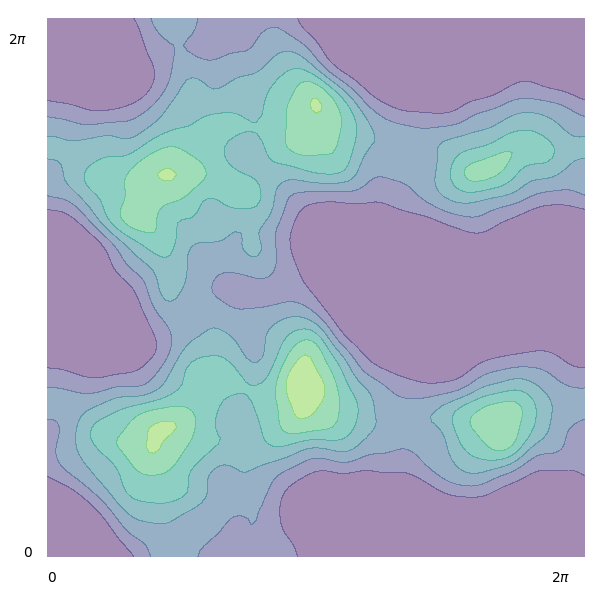}
        \caption{GFlowNet GFN2-xTB}
        \label{fig:sup:ibuprofen-gfn-xtb}
    \end{subfigure}
    \begin{subfigure}{0.25\textwidth}
        \includegraphics[width=\linewidth]{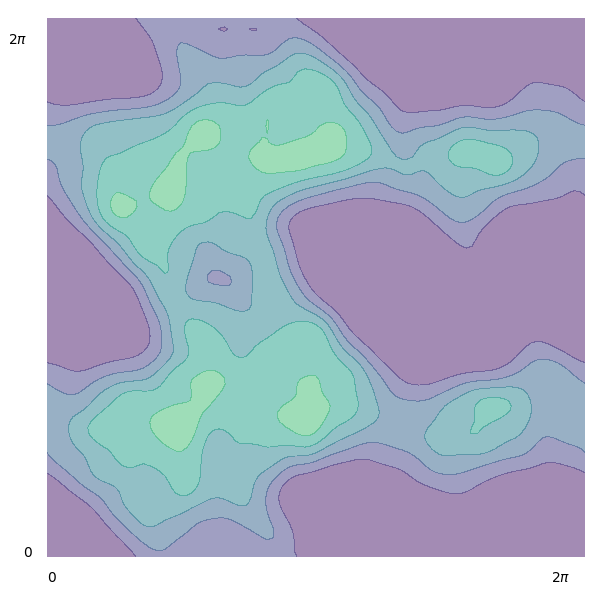}
        \caption{MCMC GFN2-xTB}
        \label{fig:sup:ibuprofen-mcmc-xtb}
    \end{subfigure}
    \\
    \begin{subfigure}{0.25\textwidth}
        \includegraphics[width=\linewidth]{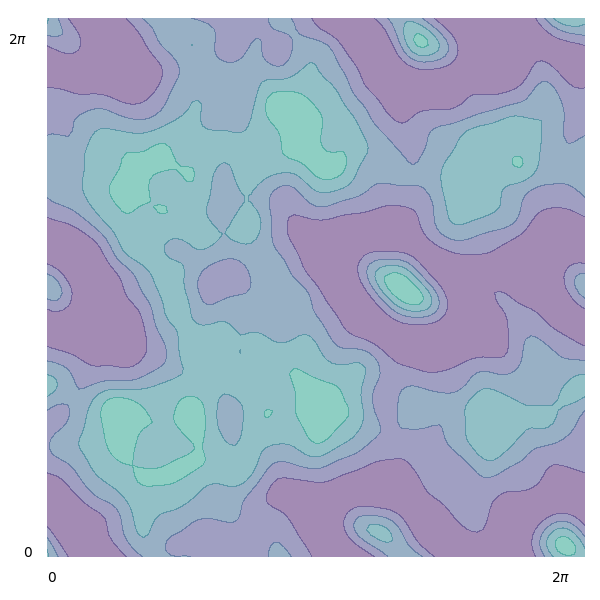}
        \caption{Reward GFN-FF}
        \label{fig:sup:ibuprofen-reward-gfn-ff}
    \end{subfigure}
    \begin{subfigure}{0.25\textwidth}
        \includegraphics[width=\linewidth]{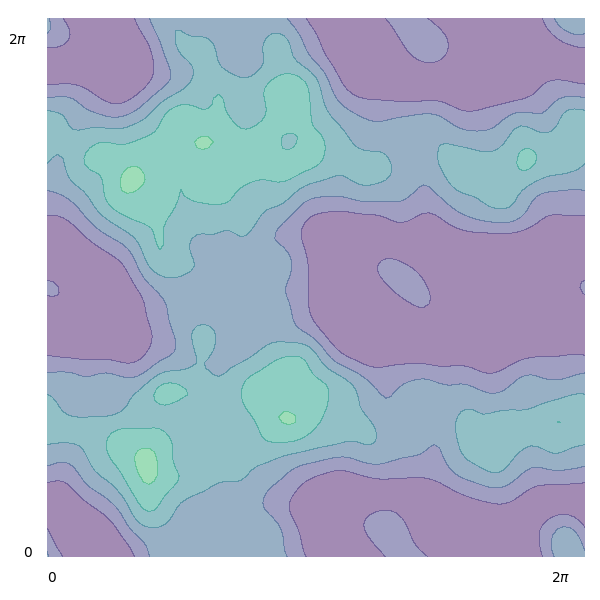}
        \caption{GFlowNet GFN-FF}
        \label{fig:sup:ibuprofen-gfn-gfn-ff}
    \end{subfigure}
    \begin{subfigure}{0.25\textwidth}
        \includegraphics[width=\linewidth]{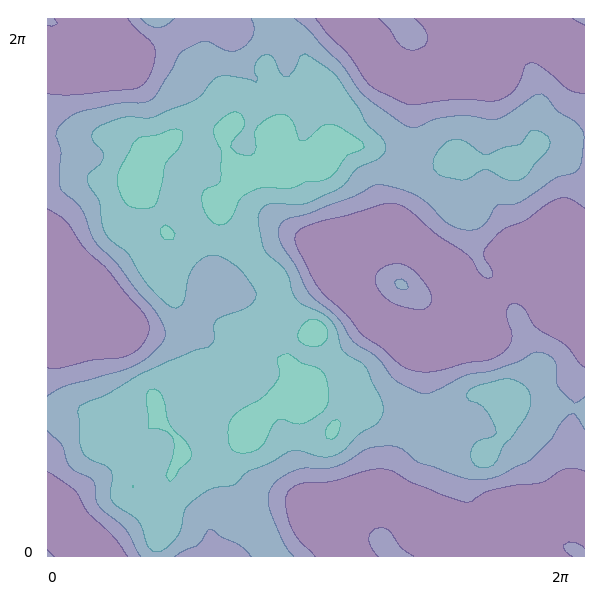}
        \caption{MCMC GFN-FF}
        \label{fig:sup:ibuprofen-mcmc-gfn-ff}
    \end{subfigure}
    \\
    \begin{subfigure}{0.25\textwidth}
        \includegraphics[width=\linewidth]{images/kde_ibuprofen_Reward_torchani.png}
        \caption{Reward TorchANI}
        \label{fig:sup:ibuprofen-reward-torchani}
    \end{subfigure}
    \begin{subfigure}{0.25\textwidth}
        \includegraphics[width=\linewidth]{images/kde_ibuprofen_GFlowNet_torchani.png}
        \caption{GFlowNet TorchANI}
        \label{fig:sup:ibuprofen-gfn-torchani}
    \end{subfigure}
    \begin{subfigure}{0.25\textwidth}
        \includegraphics[width=\linewidth]{images/kde_ibuprofen_MCMC_torchani.png}
        \caption{MCMC TorchANI}
        \label{fig:sup:ibuprofen-mcmc-torchani}
    \end{subfigure}
    \caption{KDE on samples from the reward function (left) GFlowNet (centre) and MCMC (right) for ibuprofen for the three proxies: GFN2-xTB (top), GFN-FF (middle) and TorchANI (bottom).}
\end{figure}

\clearpage
\subsection{Alanine dipeptide}

\begin{figure}[!htb]
    \centering
    \begin{subfigure}{0.25\textwidth}
        \includegraphics[width=\linewidth]{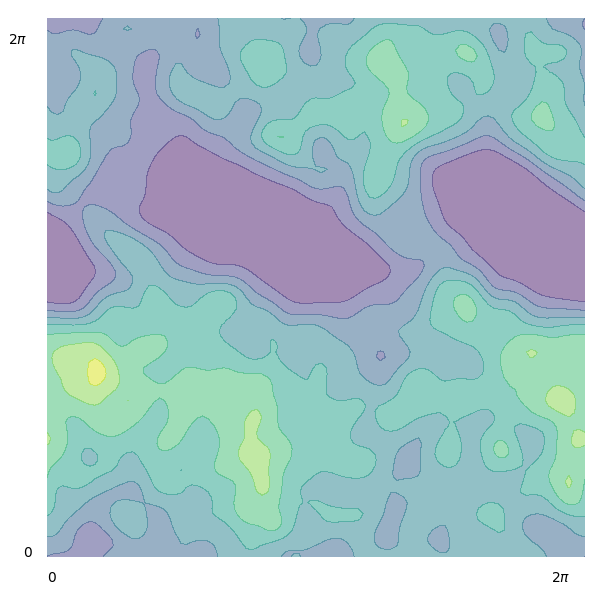}
        \caption{Reward GFN2-xTB}
        \label{fig:sup:ad-reward-xtb}
    \end{subfigure}
    \begin{subfigure}{0.25\textwidth}
        \includegraphics[width=\linewidth]{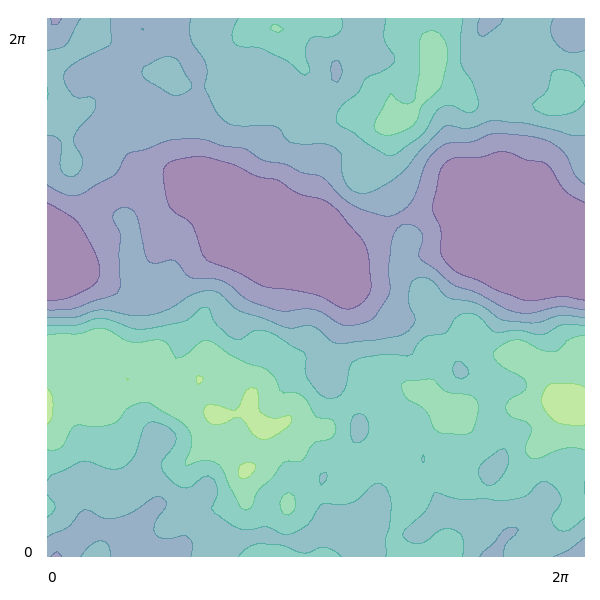}
        \caption{GFlowNet GFN2-xTB}
        \label{fig:sup:ad-gfn-xtb}
    \end{subfigure}
    \begin{subfigure}{0.25\textwidth}
        \includegraphics[width=\linewidth]{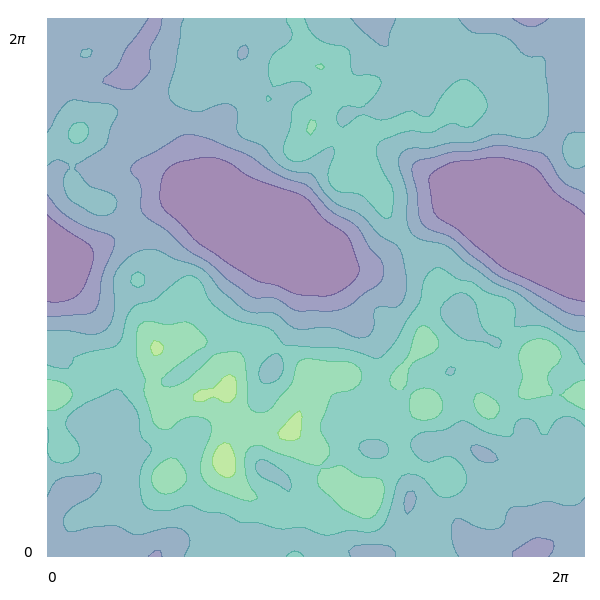}
        \caption{MCMC GFN2-xTB}
        \label{fig:sup:ad-mcmc-xtb}
    \end{subfigure}
    \\
    \begin{subfigure}{0.25\textwidth}
        \includegraphics[width=\linewidth]{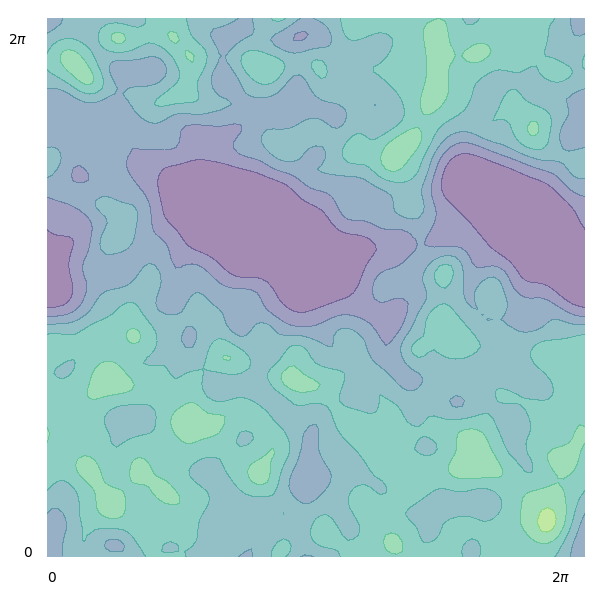}
        \caption{Reward GFN-FF}
        \label{fig:sup:ad-reward-gfn-ff}
    \end{subfigure}
    \begin{subfigure}{0.25\textwidth}
        \includegraphics[width=\linewidth]{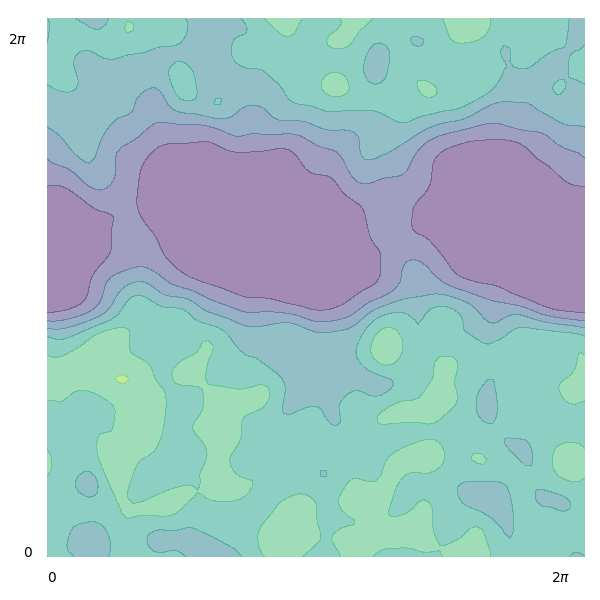}
        \caption{GFlowNet GFN-FF}
        \label{fig:sup:ad-gfn-gfn-ff}
    \end{subfigure}
    \begin{subfigure}{0.25\textwidth}
        \includegraphics[width=\linewidth]{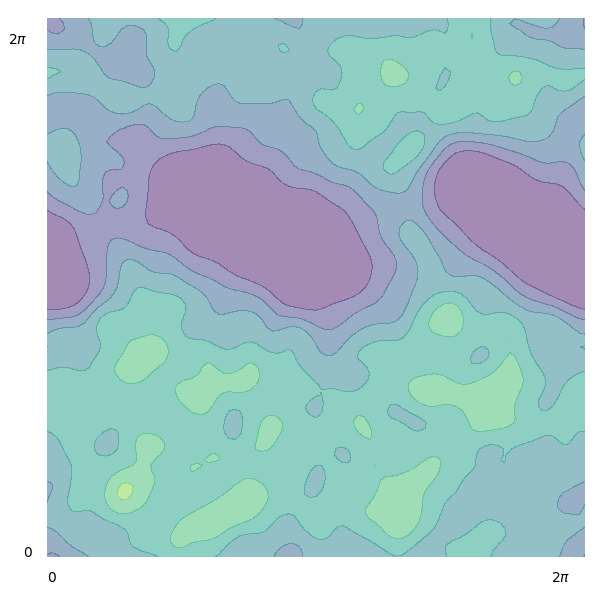}
        \caption{MCMC GFN-FF}
        \label{fig:sup:ad-mcmc-gfn-ff}
    \end{subfigure}
    \\
    \begin{subfigure}{0.25\textwidth}
        \includegraphics[width=\linewidth]{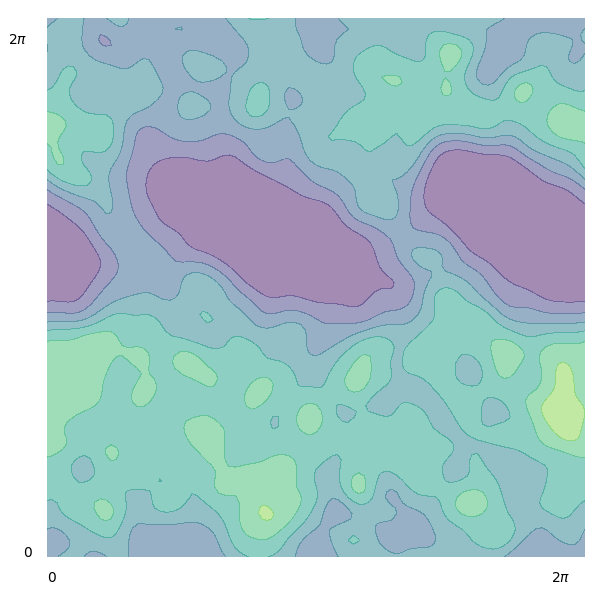}
        \caption{Reward TorchANI}
        \label{fig:sup:ad-reward-torchani}
    \end{subfigure}
    \begin{subfigure}{0.25\textwidth}
        \includegraphics[width=\linewidth]{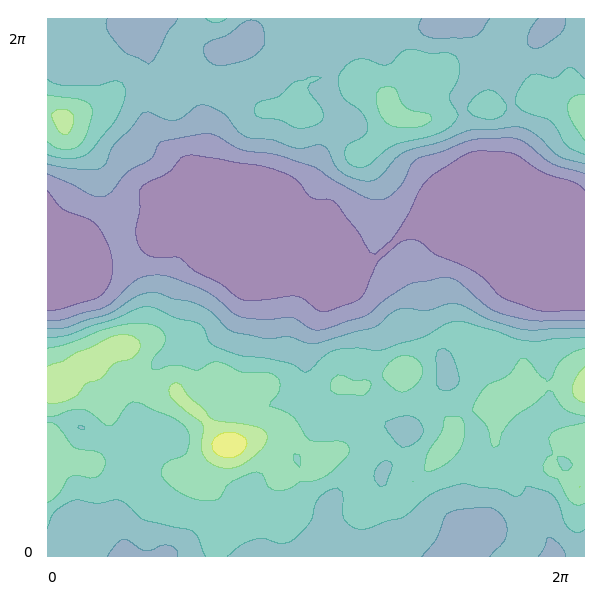}
        \caption{GFlowNet TorchANI}
        \label{fig:sup:ad-gfn-torchani}
    \end{subfigure}
    \begin{subfigure}{0.25\textwidth}
        \includegraphics[width=\linewidth]{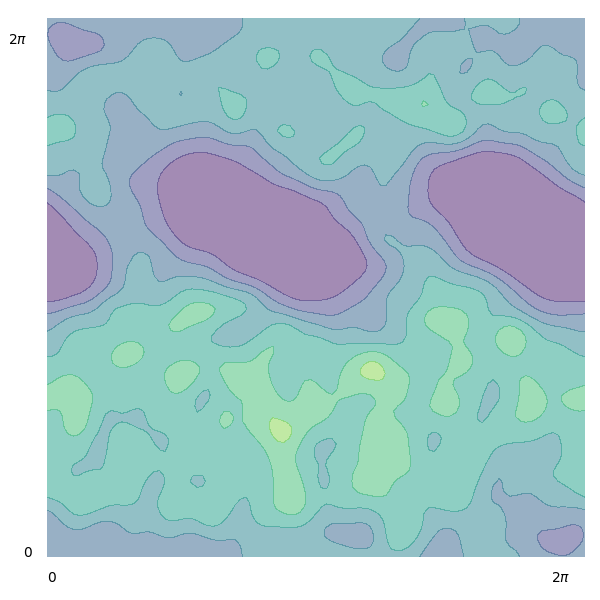}
        \caption{MCMC TorchANI}
        \label{fig:sup:ad-mcmc-torchani}
    \end{subfigure}
    \caption{KDE on samples from the reward function (left) GFlowNet (centre) and MCMC (right) for alanine dipeptide for the three proxies: GFN2-xTB (top), GFN-FF (middle) and TorchANI (bottom).}
\end{figure}








\end{document}